\documentclass[conference]{IEEEtran}
\IEEEoverridecommandlockouts
\usepackage{cite}
\usepackage{amsmath,amssymb,amsfonts}
\usepackage{algorithmic}
\usepackage{graphicx}
\usepackage{textcomp}
\usepackage{xcolor}
\usepackage{amsmath}
\usepackage{graphicx}
\usepackage{tikz}
\usetikzlibrary{shapes.geometric}
\usetikzlibrary{quotes,angles}
\usetikzlibrary{positioning}
\usepackage{caption}
\usepackage{subcaption}
\newcommand\SmallMatrix[1]{{\small\arraycolsep=\arraycolsep\ensuremath{\begin{bmatrix}#1\end{bmatrix}}}}
\newcommand{\squeezeup}{\vspace{-2mm}}
\usepackage{url}

\usepackage[left=1.91cm,right=1.91cm,top=2.54cm,bottom=1.91cm]{geometry}

\def\BibTeX{{\rm B\kern-.05em{\sc i\kern-.025em b}\kern-.08em
    T\kern-.1667em\lower.7ex\hbox{E}\kern-.125emX}}
\begin{document}
\title{Design of a Health Monitoring System for a Planetary Exploration Rover\\ 
\thanks{
This work has been accepted to the Proceedings of IEEE ISPARO 2024. \\
This document is the result of the research project funded by the Engineering and Physical Sciences Research Council. }
}

\author{\IEEEauthorblockN{Sarah Swinton}
\IEEEauthorblockA{\textit{James Watt School of Engineering} \\
\textit{University of Glasgow}\\
Glasgow, UK \\
S.Swinton.1@research.gla.ac.uk}
\and
\IEEEauthorblockN{Euan McGookin}
\IEEEauthorblockA{\textit{James Watt School of Engineering} \\
\textit{University of Glasgow}\\
Glasgow, UK \\
Euan.McGookin@glasgow.ac.uk}
\and
\IEEEauthorblockN{Douglas Thomson}
\IEEEauthorblockA{\textit{James Watt School of Engineering} \\
\textit{University of Glasgow}\\
Glasgow, UK \\
Douglas.Thomson@glasgow.ac.uk} 

}
\maketitle

\begin{abstract}
It is generally considered that a trustworthy autonomous planetary exploration rover must be able to operate safely and effectively within its environment. Central to trustworthy operation is the ability for the rover to recognise and diagnose abnormal behaviours during its operation. Failure to diagnose faulty behaviour could lead to degraded performance or an unplanned halt in operation. This work investigates a health monitoring method that can be used to improve the capabilities of a fault detection system for a planetary exploration rover. A suite of four metrics, named ‘rover vitals’, are evaluated as indicators of degradation in the rover's performance. These vitals are combined to give an overall estimate of the rover’s ‘health’. By comparing the behaviour of a faulty real system with a non-faulty observer, residuals are generated in terms of two high-level metrics: heading and velocity. Adaptive thresholds are applied to the residuals to enable the detection of faulty behaviour, where the adaptive thresholds are informed by the rover's perceived health. Simulation experiments carried out in MATLAB showed that the proposed health monitoring and fault detection methodology can detect high-risk faults in both the sensors and actuators of the rover.
\end{abstract}


\section{Introduction}

Throughout the 21st century, robotic planetary exploration missions have expanded humanity's knowledge and understanding of the solar system. These expeditions have allowed scientific investigations to be carried out in locations that are currently inaccessible to manned missions \cite{muirhead2004}.
However, all robotic systems are susceptible to faults which may degrade their performance or halt their operation and the inherent remoteness of robotic exploration missions poses a further challenge: robots cannot have hardware components repaired, replaced, or upgraded after launch. During previous NASA planetary exploration missions, fault occurrence has been recorded. The Mars Exploration Rovers (Spirit and Opportunity) experienced faults due to hardware, software, and their environment \cite{matijevic2006}. The Spirit rover experienced damage in a motor circuit that powers one of its wheels, which then had to be dragged for the remainder of the rover’s lifespan, resulting in degradation to the rovers operational performance and limitations in further mission planning \cite{mckee2006}. The Opportunity rover experienced a failure on its front-right wheel steering actuator, resulting in path following error and a reduction in precision during target approaches \cite{townsend2014}. A physical obstruction in the actuator was declared the most likely cause by an investigation team\cite{townsend2014}. These cases highlight the importance of health monitoring for planetary exploration rovers (PERs). By monitoring its own health, a PER can detect and diagnose abnormal behaviours and fault occurrence in real time. Only once these events have been identified can measures be taken to reduce any further damage or performance degradation. 

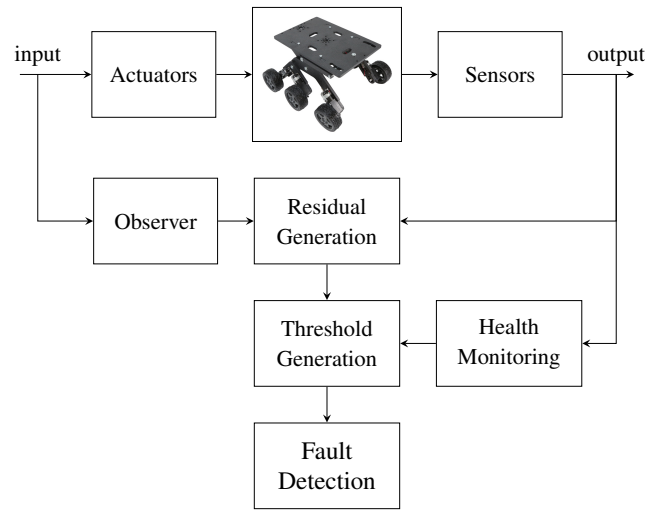
\begin{figure}[htbp]
\scalebox{0.95}{
\begin{tikzpicture}

    \node[draw,
        fill=white,
        minimum width=1.5cm,
        minimum height=1.2cm,
        text width=1.5cm, 
        align=center,
    ] (actuators) at (0,0){\small Actuators};

    \node [draw,
        fill=white,
        minimum width=2cm,
        minimum height=1.2cm,
        right=0.5cm of actuators
    ]  (rover) {{\includegraphics[width=1.85cm]{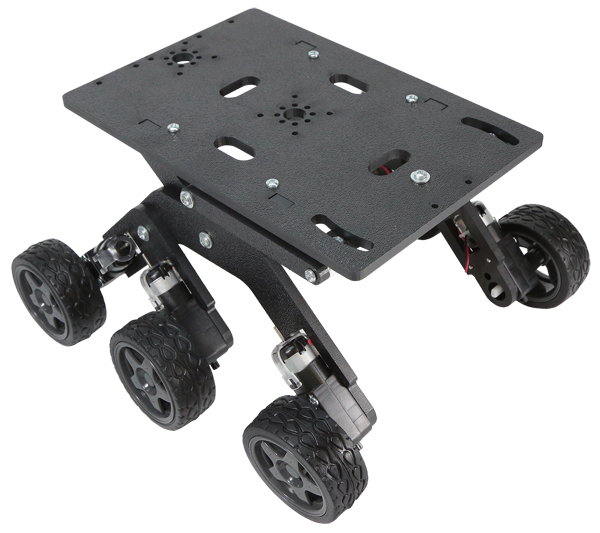}}};

    \node [draw,
        fill=white,
        minimum width=2cm,
        minimum height=1.2cm,
        anchor=south west, 
        text width=1.8cm, 
        align=center,
        below = 0.5cm of rover
    ]  (resGen) {\small  Residual \\ Generation};
    \node [draw,
        fill=white,
        minimum width=1.5cm,
        minimum height=1.2cm, 
        anchor=south west, 
        text width=1.5cm, 
        align=center,
        left = 0.5cm of resGen
    ]  (refrover) {\small Observer};

    \node [draw,
        fill=white,
        minimum width=1.5cm,
        minimum height=1.2cm,
        text width=1.5cm, 
        align=center,
        right=0.5cm of rover
    ]  (sensors) {\small Sensors};
    
    \node [draw,
        fill=white,
        minimum width=2cm,
        minimum height=1.2cm,
        anchor=south west, 
        text width=1.8cm, 
        align=center,
        below = 0.5cm of resGen
    ]  (threshGen) {\small Threshold \\ Generation};
    
    \node [draw,
        fill=white,
        minimum width=2cm,
        minimum height=1.2cm,
        anchor=south west, 
        text width=1.8cm, 
        align=center,
        right = 0.5cm of threshGen 
    ]  (healthMonitor) {\small Health \\ Monitoring};
    
    \node [draw,
        fill=white,
        minimum width=2cm,
        minimum height=1.2cm,
        anchor=south west, 
        text width=1.8cm, 
        align=center,
        below = 0.5cm of threshGen
    ]  (fd) {Fault \\ Detection};

    \node[left = 1cm of actuators
    ]  (pointA) {};    
    \node[right = 1cm of sensors
    ]  (pointB) {};

    \draw[-stealth] (pointA) -- (actuators.west)
        node[near start,above](input){\small input};
    \draw[-stealth] (input) |- (refrover.west) 
        node[near end,left]{};
    \draw[-stealth] (actuators.east) -- (rover.west) 
        node[midway,above]{};
    \draw[-stealth] (rover.east) -- (sensors.west) 
        node[midway,above]{};
    \draw[-stealth] (refrover.east) -- (resGen.west) 
        node[midway,above]{};
    \draw[-stealth] (resGen.south) -- (threshGen.north) 
        node[midway,above]{};
    \draw[-stealth] (threshGen.south) -- (fd.north) 
        node[midway,above]{};
    \draw[-stealth] (healthMonitor.west) -- (threshGen.east) 
        node[midway,above]{};
    \draw[-stealth] (sensors.east) -- (pointB)
        node[near end,above](output){\small output};
    \draw[-stealth] (output) |- (resGen.east) 
        node[near end,left]{};
    \draw[-stealth] (output) |- (healthMonitor.east) 
        node[near end,left]{};


\end{tikzpicture}}
\caption{PER fault detection system architecture}
\label{fig:FDArchitecture}
\end{figure}

Fig. \ref{fig:FDArchitecture} shows the architecture of a PER fault detection system. The rover's control input is given to both a non-faulty reference model, which acts as an observer, and the faulty rover, which represents the real system and consists of actuators, rover kinematic and dynamic models, and sensors. The outputs of the reference model and faulty rover are compared to generate output error residuals. By considering the rover's health, adaptive thresholds are generated to allow the detection of faulty behaviour.

Ramesh et al set out a method for robot health monitoring using `robot vitals’ \cite{ramesh2021}. These vitals are parameters that measure the robot’s performance, and in doing so, can capture performance degradation in real-time. By generating a suite of robot vitals that fully encapsulate the behaviours of a robot, the overall health of the robot can be measured.  

The contributions of this work are: (a) a novel implementation of the robot vitals framework proposed by Ramesh (\cite{ramesh2021}, \cite{ramesh2022}) for PERs and (b) the extension of this method to inform the design of fault detection systems using adaptive thresholds.

This work is set out as follows. Section \ref{sec:systemModel} sets out the mathematical model of a small PER subject to faults. Section \ref{sec:roverVitals} defines a set of rover vitals to quantify the probability of a rover experiencing performance degradation due to faults at a given time. Section \ref{sec:roverHealth} presents a methodology for measuring overall rover health by observing rover vitals. Section \ref{sec:faultDetection} proposes a methodology for fault detection whereby rover health informs adaptive thresholds on high-level output error residuals. 
Section \ref{sec:results} details the simulation experiments carried out to evaluate the proposed health monitoring strategy. A discussion of these results is presented in Section \ref{sec:discussion}. Finally, Section \ref{sec:conclusions} summarises the conclusions that can be drawn from this work.

\section{System Modelling} \label{sec:systemModel}
\subsection{PER Model} \label{subsec:rovermodel}
\begin{figure} [h]
\centering
\scalebox{1.1}{
\begin{tikzpicture} 
    
    \node[inner sep=0pt] (rover) at (-2,1.6)
    {\includegraphics[width=.08\textwidth]{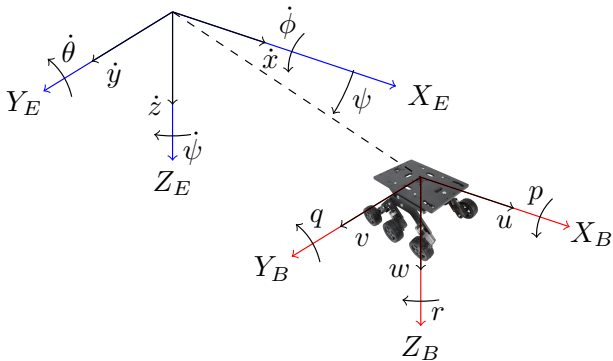}};

    \draw (-5,4) coordinate (oe) node[left = 5pt, below=1pt] {};
     \draw[draw, blue, ->] (-5,4) -- ++(0,-1.5*1.2) node[pos=1.15, yshift=0mm, xshift=0mm, black]{$Z_E$};
    \draw[draw, blue, ->] (-5,4) -- ++(1.5*1.8,-0.5*1.8) node[pos=1.15, yshift=0mm, xshift=0mm, black]{$X_E$};
    \draw[draw, blue, ->] (-5,4) -- ++(-1.3*1.2,-0.8*1.2) node[pos=1.15, yshift=0mm, xshift=0mm, black]{$Y_E$};

    \draw (-2,2) coordinate (ob) node[left = 5pt, below=1pt] {};
    \draw[draw, red, ->] (-2,2) -- ++(0,-1.5*1.2) node[pos=1.15, yshift=0mm, xshift=0mm, black]{$Z_B$};
    \draw[draw, red, ->] (-2,2) -- ++(1.5*1.2,-0.5*1.2) node[pos=1.15, yshift=0mm, xshift=0mm, black]{$X_B$};
    \draw[draw, red, ->] (-2,2) -- ++(-1.3*1.2,-0.8*1.2)node[pos=1.15, yshift=0mm, xshift=0mm, black]{$Y_B$};

    \draw[dashed] (oe) -- (ob);   

    \draw[draw, ->] (-5,4) -- ++(0,-1.5*0.75)  node [left] {$\dot{z}$};
    \draw[draw, ->] (-5,4) -- ++(1.5*0.75,-0.5*0.75) node[pos=0.9, yshift=-2.2mm, xshift=1.7mm, black]{$\dot{x}$}; 
    \draw[draw, ->] (-5,4) -- ++(-1.3*0.75,-0.8*0.75) node [pos=0.9, yshift=-2.2mm, xshift=1.7mm, black] {$\dot{y}$};

    \draw[draw, ->] (-2,2) -- ++(0,-1.5*0.75) node [left] {$w$};
    \draw[draw, ->] (-2,2) -- ++(1.5*0.75,-0.5*0.75) node[pos=0.9, yshift=-2mm, xshift=0mm, black]{$u$}; 
    \draw[draw, ->] (-2,2) -- ++(-1.3*0.75,-0.8*0.75) node [pos=0.9, yshift=-2.2mm, xshift=1.7mm, black] {$v$};

    \node (a) at (-4.85,3) {};
    \node (b) at (-5.15,3) {};
    \node (c) at (-5,) {};
    \pic [draw, <-, angle radius=15mm, angle eccentricity=1.1,right, "$\dot{\psi}$"] {angle = b--oe--a};

    \node (e) at (-3.6,3.95) {};
    \node (f) at (-3.9,3.25) {};
    \node (efo) at (-3,3.3) {};
    \pic [draw, ->, angle radius=6mm, angle eccentricity=1.1, above, "$\dot{\phi}$"] {angle = e--efo--f};

    \node (i) at (-5.2,3.2) {};
    \node (j) at (-6.3,3.7) {};
    \node (ijo) at (-7,2.8) {};
    \pic [draw, ->, angle radius=8mm, angle eccentricity=1.1, above, "$\dot{\theta}$"] {angle = i--ijo--j};

    \node (c) at (-1.85,1) {};
    \node (d) at (-2.15,1) {};
    \pic [draw, <-, angle radius=15mm, angle eccentricity=1.1, right, "$r$"] {angle = d--ob--c};
    
    \node (g) at (-0.6,1.95) {};
    \node (h) at (-0.9,1.25) {};
    \node (gho) at (0,1.3) {};
    \pic [draw, ->, angle radius=6mm, angle eccentricity=1.1, above, "$p$"] {angle = g--gho--h};
    
    \node (k) at (-2.2,1.2) {};
    \node (l) at (-3.3,1.7) {};
    \node (klo) at (-4,0.8) {};
    \pic [draw, ->, angle radius=8mm, angle eccentricity=1.1, above, "$q$"] {angle = k--klo--l};
    
    \node (m) at (-3.6,3.55) {};
    \node (n) at (-3.9,3.3) {};
    \pic [draw, <-, angle radius=23mm, angle eccentricity=1.1, "$\psi$"] {angle = n--oe--m};

\end{tikzpicture}}
\caption{Earth-fixed (X\textsubscript{E}, Y\textsubscript{E}, Z\textsubscript{E} (blue)) and rover body-fixed axes (X\textsubscript{B}, Y\textsubscript{B}, Z\textsubscript{B} (red)) for the modelled rover}
\label{fig:framesOfRef}
\end{figure}

In this work, a small six-wheel rover with rocker-bogie suspension and differential steering is used. Two frames of references are utilised (Fig. \ref{fig:framesOfRef}). The first is the Earth-fixed frame, which has an inertially fixed origin, and axes denoted X\textsubscript{E}, Y\textsubscript{E}, Z\textsubscript{E}. The second is the rover body frame, which rotates with the rover's motion i.e., is fixed to the rover’s axes denoted X\textsubscript{B}, Y\textsubscript{B}, Z\textsubscript{B}. The origin of the rover body frame is the rover’s centre of gravity. These axes are oriented following the North East Down (NED) system, where positive Z motion is directed downwards from the rover's centre of gravity \cite{fossen1994}.  
\squeezeup

\squeezeup Here, the rover's linear velocities are $u$, $v$, and $w$. The rover's rotational velocities are $p$, $q$, and $r$. 
The rover's equations of motion, with reference to the rover's body-fixed frame and Earth-fixed frame, can be described by the matrix relationships shown in Equations (\ref{eqn:eqOfMotion}, \ref{eqn:eqOfMotionComp1}, \ref{eqn:eqOfMotionComp2}) \cite{fossen1994}. 

\renewcommand\arraystretch{1.5}
\begin{equation}
    \SmallMatrix{\boldsymbol{\dot{v}}\\\boldsymbol{\dot{\eta}}}
  = 
  \SmallMatrix{\boldsymbol{\alpha}(v) & \boldsymbol{\beta}(v)\\ \boldsymbol{J}(\eta) & 0}
  \SmallMatrix{\boldsymbol{v} \\ \boldsymbol{\eta}}
  +
  \SmallMatrix{ -\boldsymbol{M}^{-1} \\0 } \boldsymbol{\tau}
  \label{eqn:eqOfMotion}
\end{equation} 
\begin{equation}
    \boldsymbol{\alpha} (v)= -(\boldsymbol{C}(v)+\boldsymbol{D}(v)) \boldsymbol{M}^{-1}
  \label{eqn:eqOfMotionComp1}
\end{equation}
\begin{equation}
    \boldsymbol{\beta} (v) = -\boldsymbol{g}(v) \boldsymbol{M}^{-1}
  \label{eqn:eqOfMotionComp2}
\end{equation}

Here, $\boldsymbol{v}$ is the body-fixed velocity vector and $\boldsymbol{\eta}$ is the inertially fixed position/orientation vector. $\boldsymbol{M}$ is the mass and inertia matrix, $\boldsymbol{C}(v)$ is the Coriolis  matrix, $\boldsymbol{D}(v)$ is the damping  matrix, $\boldsymbol{g}(v)$ represents  the gravitational  forces  and  moments, $\boldsymbol{J}(\eta)$ is an Euler matrix representing the trigonometric transformation from the body fixed reference frame to the earth fixed reference frame, and the $\boldsymbol{\tau}$ vector  represents the  forces and moments generated by the actuators. The actuators are provided with commands as set out in Section \ref{subsec:gnc} . 

\subsection{Guidance, Navigation and Control} \label{subsec:gnc}

A line-of sight navigation algorithm \cite{breivik2003} is utilised, which allows the rover to navigate towards its current target. A target point is classed as `collected' if the centre point of the rover is within the acceptance radius, i.e. one half of the rover's length to provide sufficient navigation accuracy for this application. The control system consists of two PID controllers for heading and velocity, respectively. Due to the closed loop nature of the control system, the rover can compensate for some changes in behaviour due to both external factors (e.g. interaction with the environment) and internal factors (e.g. non-critical internal rover faults). As the control may adapt in real time to some faulty behaviours, it is important to be able to monitor the overall health of the rover to prevent any further damage or performance degradation.

\subsection{Fault Modelling}
For this work, two high-risk internal fault modes are evaluated, one on the IMU (gyroscope offset) and one on the actuators (motor failure).  Faults in the sensors and actuators of a rover are unlikely to cause a halt in the rover’s operation, and instead would result in a degradation to the rover’s performance \cite{swinton2022fault}\cite{aloufi2022}. This performance degradation can be captured by monitoring the health of the rover.

\subsubsection{Gyroscope Offset}
Gyroscope offset can be caused by damage to the IMU. In this work, a stepwise offset has been applied to represent a fault in the measurement of the PER heading output. 

\subsubsection{Motor Failure}
Motor failures have been reported during planetary exploration missions, due to both adverse terrain interactions \cite{rankin2020} and internal circuitry failures \cite{townsend2014}. It is therefore important to consider the effect of motor failures on the health of a rover. In this work, an internal motor failure has been considered, whereby the motor driving the rover’s front left wheel fails to rotate. The wheel therefore drags along the surface, instead of generating propulsive force. This models a real-world scenario where the circuitry powering the motor has been damaged.

\squeezeup
\section{Rover Vitals}\label{sec:roverVitals}
\subsection{Introduction}
As discussed in the work by Ramesh, \textit{vitals} are measures of a robot’s real-time performance degradation during operation \cite{ramesh2022}. Vitals should be selected that are both task agnostic and correlated to performance degradation \cite{ramesh2021}. No single vital can give a full picture of a robot’s overall health.  By defining an appropriate suite of vitals, the performance of a robot can be sufficiently captured and used to diagnose behavioural abnormalities. 
Ramesh et al show that a suite of vitals is effective in highlighting performance degradation in a single, self-monitoring robot \cite{ramesh2022}. 
 
To define an appropriate suite of vitals for a planetary rover system, the nominal behaviours of the rover must be observed. Each vital is designed to output values in the range $[0,1]$ to indicate the likelihood of performance degradation within the system. For each vital, nominal behaviour indicates that no performance degradation is present, therefore the vital should output a value close to $0$. Conversely, abnormal behaviours indicate likely performance degradation, and as such the vital should output a value close to $1$. 

In this work, four vitals are defined in order to capture the PER’s overall performance: \textit{forward acceleration}, \textit{rate of change of distance to goal}, \textit{rate of change of heading angle}, and \textit{rate of change of commanded motor voltage}. The following subsections set out the probability distribution functions for each vital, which have been designed with reference to the nominal behaviours of the PER. For vitals which utilise a Gaussian distribution, a value of $\sigma = 0.4$ has been selected. This ensures that ideal rover behaviour provides an almost zero output for the given vital.

\subsection{Forward Acceleration}
Many high-risk fault modes could cause an abrupt spike in a rover’s forward acceleration. These include, though are not limited to, a loss of power resulting in a halt, a failure within one or more actuators causing a change in propulsive force, and accelerometer offsets within the inertial measurement unit (IMU). Under nominal conditions, the rover’s acceleration should remain low (or close to zero) during path traversal. It is expected that a spike in acceleration will occur whenever the rover collects its current target and turns towards its next target. Within this work, each rover must collect only one target, and so only an initial spike is expected. The acceleration vital function, $\boldsymbol{V}(a_x)$, is calculated using a Gaussian distribution: 

\begin{equation}
    \boldsymbol{V}(a_x) = 1 - \dfrac{1}{\sigma \sqrt{2\pi}} e^{-\dfrac{a_x^2}{2\sigma^2}}
\end{equation} \newline
Here $\sigma = 0.4$. This value  of $\sigma$ has been selected in line with an analysis of the nominal behaviour of the rover, to ensure non-faulty rover performance provides a low value of $\boldsymbol{V}(a_x)$.

\subsection{Rate of Change of Distance to Target}
Within this work, rovers are tasked with collecting a target point. A straight-line path between the rover’s start point and target point has been assumed. The rate of change of distance to target, $\dot d_t$, is therefore measured using the Euclidean distance from the rover to its target point at each time step. Under nominal conditions, $\dot d_t$ should remain steady at a negative value as the rover moves towards its goal. However, the presence of faults or abnormal behaviour can affect  $\dot d_t$. For example, a halted rover will experience  $\dot d_t$ approximately equal to zero, and a rover experiencing faults may be unable to follow its planned path, resulting in  $\dot d_t$ greater than or equal to zero. The  $\dot d_t$ vital function, $\boldsymbol{V}(\dot d_t)$, is modelled using a logistic sigmoid function: 
\begin{equation}
    \label{eqn:vital_dot_dt}
    \boldsymbol{V}(\dot d_t) = \dfrac{1}{1+e^{-k(\dot d_t-x_0)}} 
\end{equation} 
Here $k = 20$ and $x_0 = 0.1$.  Equation \ref{eqn:vital_dot_dt} is designed such that a rover travelling at the nominal forward velocity $v_x = 0.25 m/s$ gives a low value of $\boldsymbol{V}(\dot d_t)$.

\subsection{Rate of Change of Heading Angle}
A rover should turn towards its target point during the initial stages of path traversal. Therefore, non-zero values of heading rate are to be expected. However, once the rover is pointing at its target, it should not deviate from its straight path. Various high risk fault modes, such as increased actuator friction, and constant output bias offset in the IMU \cite{aloufi2022} could cause changes to the heading of the rover. The closed loop control system implemented for the rovers may be able to compensate for the effects of such faults, but in doing so a spike in heading rate should be observed. The heading rate vital function, $\boldsymbol{V}(\dot \psi)$, is calculated using a Gaussian distribution: 
\begin{equation}
    \boldsymbol{V}(\dot \psi) = 1 - \dfrac{1}{\sigma \sqrt{2\pi}} e^{-\dfrac{\dot \psi^2}{2\sigma^2}}
\end{equation} \newline
Here $\sigma = 0.4$. This value  of $\sigma$ has been selected in line with an analysis of the nominal behaviour of the rover, to ensure non-faulty rover performance provides a low value of $\boldsymbol{V}(\dot \psi)$.

\subsection{Rate of Change of Commanded Voltage}
The rover considered in this work has a closed loop control system. This can compensate for some faulty behaviours, such as a single halted wheel, or increased friction in the rover's motors, by changing the commanded voltage, $V_c$ sent to the motors. The commanded voltage vital function, $\boldsymbol{V}(V_c)$, is calculated using a Gaussian distribution: 

\begin{equation}
    \boldsymbol{V}(V_c) = 1 - \dfrac{1}{\sigma \sqrt{2\pi}} e^{-\dfrac{V_c^2}{2\sigma^2}}
\end{equation}\newline
Here $\sigma = 0.4$. This value  of $\sigma$ has been selected in line with an analysis of the nominal behaviour of the rover, to ensure non-faulty rover performance provides a low value of $\boldsymbol{V}(V_c)$.

\section{Rover Health}\label{sec:roverHealth}
Rover health is a scalar value that describes the rover’s ability to carry out its tasks \cite{ramesh2022}. By monitoring this value, abnormal behaviour can be identified. To obtain a measure of rover health, the overall likelihood of performance degradation must first be evaluated using the suite of vitals generated in Section \ref{sec:roverVitals}.
At time $t$, the suite of vitals is given by the vector $\boldsymbol{V}_t$, as shown in Equation (\ref{eqn:vitalVector}).

\begin{equation}
    \boldsymbol{V}_t = \{a_x,\dot d_t, \dot \psi,V_c\}
    \label{eqn:vitalVector}
\end{equation}

At a given time step, each component of $V_t$ is summed, as shown in Equation (\ref{eqn:probSufferingTotal}), where $\eta$ is a normalisation factor (i.e. $\eta = 0.25$), to give the probability of performance degradation, $P$.
\begin{equation}
    P = \eta \sum_{i=1}^{4} V_{i}|_t
    \label{eqn:probSufferingTotal}
\end{equation} 

\indent Information Entropy, or Shannon Entropy, was first introduced to measure the uncertainty, or `surprise', associated with the possible outcomes of a variable \cite{shannon1948}. To obtain a measure of the rover's overall  health, the information entropy associated with the potential outcomes is evaluated. In this work there are two potential outcomes: faulty or non-faulty. If the likelihood of both potential outcomes is equal, information entropy, $H$, is 1. If the difference in the two probabilities increases, $H$, decreases. This means that the information entropy will be low in cases where it is likely that no fault is present. 
The health of the rover is then evaluated using Equation (\ref{eqn:health}) such that  $H \approx1$ indicates little to no performance degradation. A second order filter is used to reduce the effects of noise on $H$.
\begin{equation}
    H = 1-\sum P(x) \log\left(\dfrac{1}{P(x)}\right)
    \label{eqn:health}
\end{equation} 
As this measure of health considers instantaneous performance degradation, if a system with a closed loop control system (such as the PER considered in this work) is able to compensate for faulty behaviour online, the health value will return to pre-fault levels. 

\section{Fault Detection}\label{sec:faultDetection}

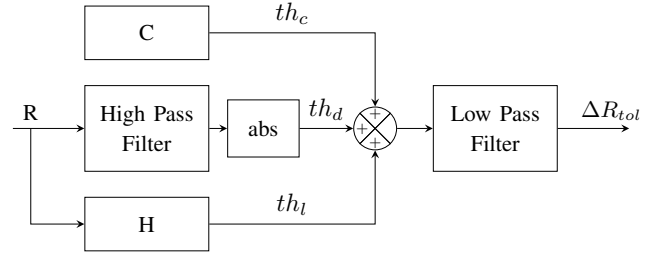
\begin{figure}[htbp]
\scalebox{0.95}{
\begin{tikzpicture}

    \node[draw,
        fill=white,
        minimum width=1.5cm,
        minimum height=1.2cm,
        text width=1.5cm, 
        align=center,
    ] (highPassFilter) at (0,0){\small High Pass\\ Filter};

    \node [draw,
        fill=white,
        minimum width=1cm,
        minimum height=0.8cm,
        right=0.25cm of highPassFilter
    ]  (absVal) {\small abs};
    
    \node [draw,
        fill=white,
        minimum width=1cm,
        minimum height=0.75cm,
        anchor=south west, 
        text width=1.5cm, 
        align=center,
        below = 0.35cm of highPassFilter
    ]  (health) {\small H};
    
    \node[draw,
        circle,
        minimum size=0.6cm,
        fill=white,
        right = 0.75cm of absVal
    ] (sum){};
     
    \draw (sum.north east) -- (sum.south west)
        (sum.north west) -- (sum.south east);
     
    \draw (sum.north east) -- (sum.south west)
    (sum.north west) -- (sum.south east);
     
    \node[left=-1pt] at (sum.center){\tiny $+$};
    \node[below] at (sum.center){\tiny $+$};
    \node[above] at (sum.center){\tiny $+$};
    
    \node [draw,
        fill=white,
        minimum width=1cm,
        minimum height=0.75cm,
        anchor=south west, 
        text width=1.5cm, 
        align=center,
        above = 0.35cm of highPassFilter
    ]  (c) {\small C};
    
    \node [draw,
        fill=white,
        minimum width=1.2cm,
        minimum height=1.2cm,
        anchor=south west, 
        text width=1.5cm, 
        align=center,
        right=0.5cm of sum
    ]  (lowPassFilter){\small Low Pass\\ Filter};

    \node[left = 1cm of highPassFilter
    ]  (pointA) {};    
    \node[right = 1cm of lowPassFilter
    ]  (pointB) {};

    \draw[-stealth] (pointA) -- (highPassFilter.west)
        node[near start,above](input){\small R};
    \draw[-stealth] (input) |- (health.west) 
        node[near end,left]{};
    \draw[-stealth] (highPassFilter.east) -- (absVal.west) 
        node[midway,above]{};
    \draw[-stealth] (absVal.east) -- (sum.west) 
        node[midway,above]{$th_{d}$};
    \draw[-stealth] (health.east) -| (sum.south) 
        node[near start,above]{$th_{l}$};
    \draw[-stealth] (c.east) -| (sum.north) 
        node[near start,above]{$th_c$};
    \draw[-stealth] (sum.east) -- (lowPassFilter.west) 
        node[midway,above]{};
    \draw[-stealth] (lowPassFilter.east) -- (pointB)
        node[near end,above](input){\small $\Delta R_{tol} $}; 
\end{tikzpicture}}
\caption{Adaptive threshold block diagram}
\label{fig:AdaptiveThresholdArchitecture}
\end{figure}

The fault detection system implemented in this work follows the architecture presented in Fig. \ref{fig:FDArchitecture}, where the outputs of the reference model and faulty rover are compared to generate output error residuals in terms of both heading, $R_{\psi}$, and velocity, $R_V$. 
Fault detection is carried out by placing thresholds on the output error residuals. Adaptive thresholds are used in fault detection in order to reduce the number of false positive detections. In model-based fault detection systems, these false positive detections may occur in cases where there are inaccuracies in the model, or where excessive noise is present.
By monitoring system parameters to dynamically adjust fault detection thresholds, changes in the system can be identified. 
Fig. \ref{fig:AdaptiveThresholdArchitecture} shows the design of the adaptive threshold applied to the generated residuals.

\vspace*{-5mm}
Here $R$ is the input to the threshold generator (i.e. $R_{\psi}$ or $R_V$). Three adaptive threshold components are generated. The first is $th_c$, which provides a constant component, $c$, tuned separately for the respective residuals. The second is $th_d$ which provides a dynamic component, whereby the input, $U$, is passed through a high pass filter to encapsulate the the dynamic range of the input signal, and the absolute value is taken. The third component is, $th_{l}$. This provides a linear component, where the input, $R$, is scaled in terms of the rover's measured health. The components $th_c$, $th_d$, and $th_{lin}$ are the summed, and a low pass filter is used to smooth the resulting value of $\Delta R_{tol}$ i.e. the maximum acceptable change in residual.

\squeezeup
\section{Results}\label{sec:results}

\subsection{Experimental Set-Up}
A MATLAB simulation has been developed to allow testing of the proposed methods. In this simulation, two scenarios have been created, in each of which a single rover (as defined in Section \ref{sec:systemModel}) must collect a set of waypoints by traversing a 3D surface, which has been generated using HiRISE digital terrain modelling of Jezero crater \cite{HiRISE}. In the first scenario, the rover must travel in a straight line towards a target point (Fig. \ref{fig:1a}). In the second scenario, the rover must follow a serpentine path (Fig. \ref{fig:2a}).
In each scenario, three separate test cases have been examined, in which the fault mode has been varied: 
\begin{itemize}
    \item Test Case A: No fault present.
    \item Test Case B: Gyroscope offset injected at $t = 5$ seconds such that the IMU heading output is offset by $10^{\circ}$. 
    \item Test Case C: Motor failure injected at $t = 5 $ seconds such that the front left wheel can no longer generate propulsive force and drags along the surface. 
\end{itemize}
For each scenario and test case, the rover health signals are recorded, alongside the residuals ($R_{\psi}$ and $R_V$) with the generated adaptive threshold, and a static threshold for comparison. 
\vspace*{-2mm}
\subsection{Straight Line Path}
\begin{figure*}[t]
    \begin{subfigure}[t]{0.32\textwidth}
        \includegraphics[width=\linewidth]{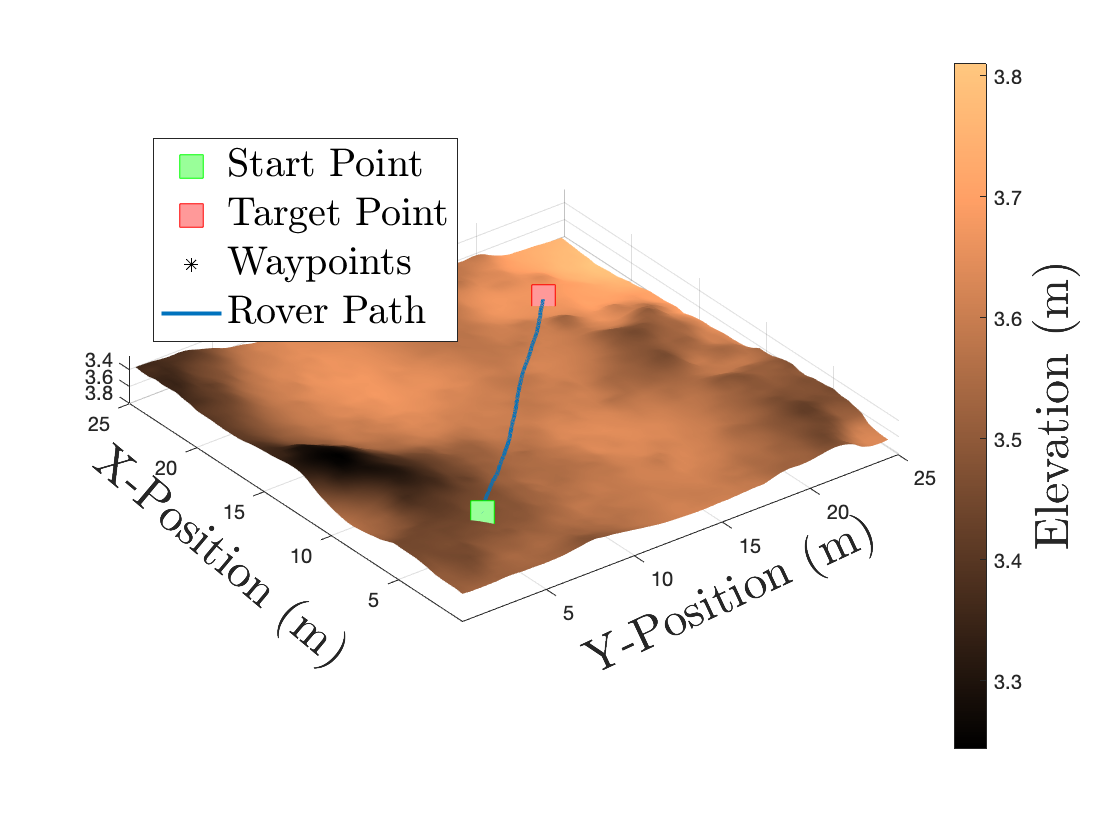}
        \caption{Nominal rover path in the presence of no faults} \label{fig:1a}
    \end{subfigure}\hspace*{\fill}
    \begin{subfigure}[t]{0.32\textwidth}
        \includegraphics[width=\linewidth]{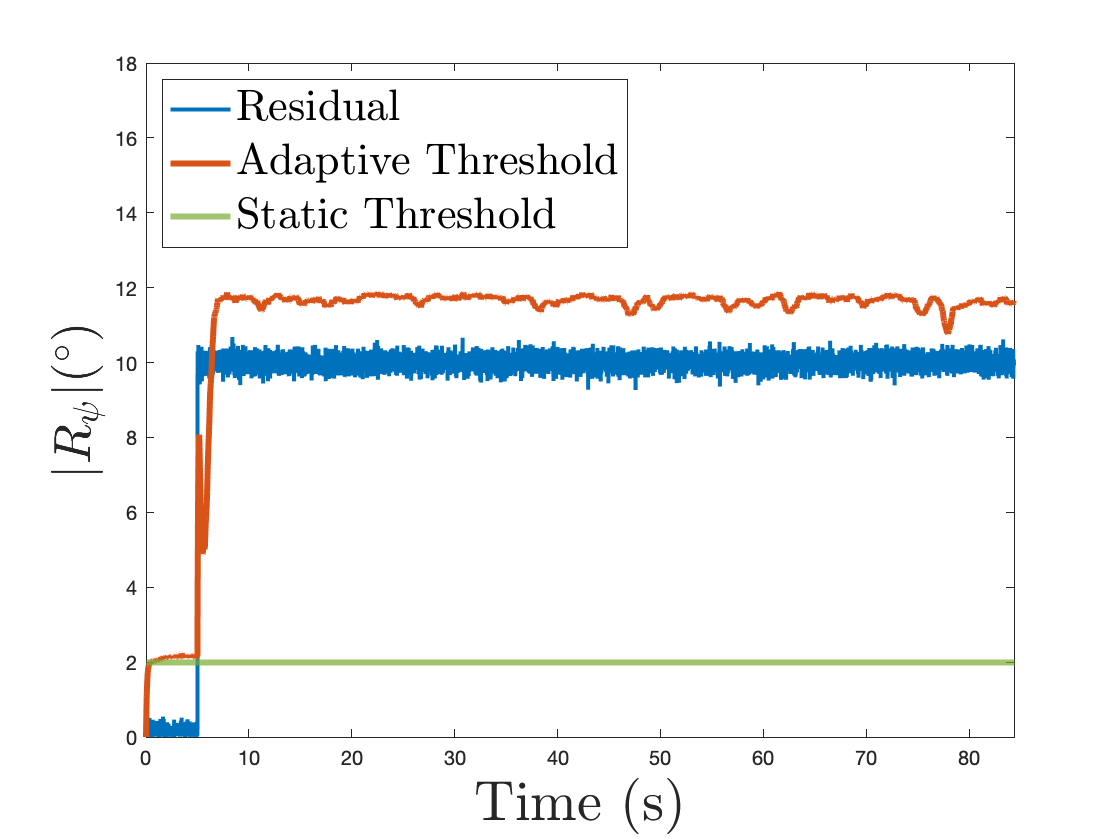}
        \caption{Test case B: heading residual with adaptive threshold in the presence of an abrupt gyroscope offset at $t=5$ seconds} \label{fig:1b}
    \end{subfigure} \hspace*{\fill}
    \begin{subfigure}[t]{0.32\textwidth}
        \includegraphics[width=\linewidth]{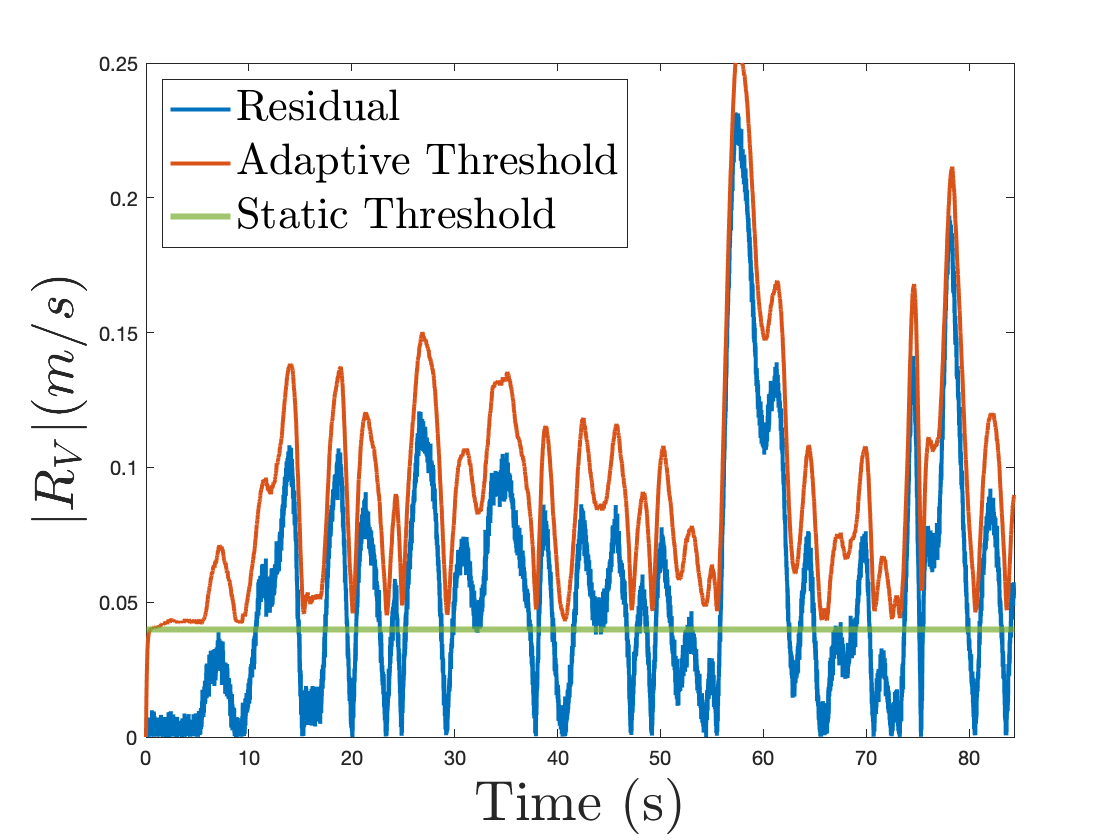}
        \caption{Test case B: velocity residual with adaptive threshold in the presence of an abrupt gyroscope offset at $t=5$ second} \label{fig:1c}     
    \end{subfigure}

    \vspace*{-2mm}

    \medskip
    \begin{subfigure}[t]{0.32\textwidth}
        \includegraphics[width=\linewidth]{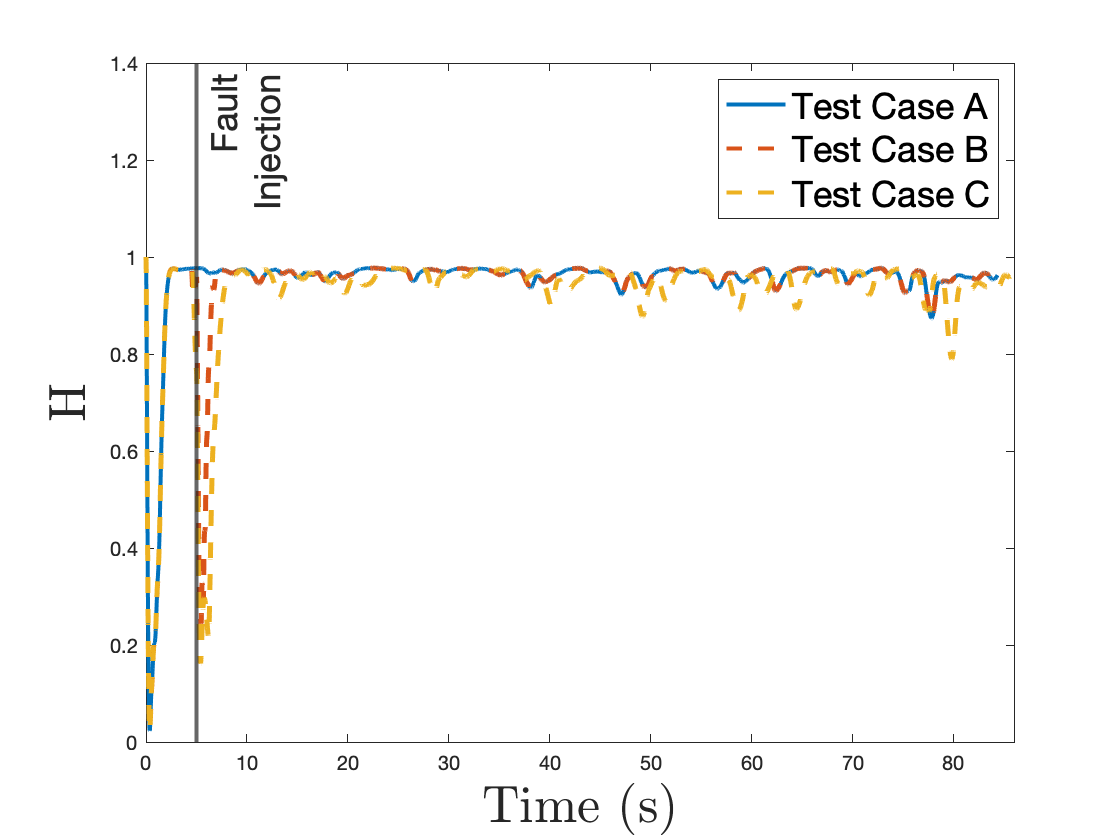}
        \caption{A comparison of the rover health signals for each of the three fault test cases} \label{fig:1d}
    \end{subfigure}\hspace*{\fill}
    \begin{subfigure}[t]{0.32\textwidth}
        \includegraphics[width=\linewidth]{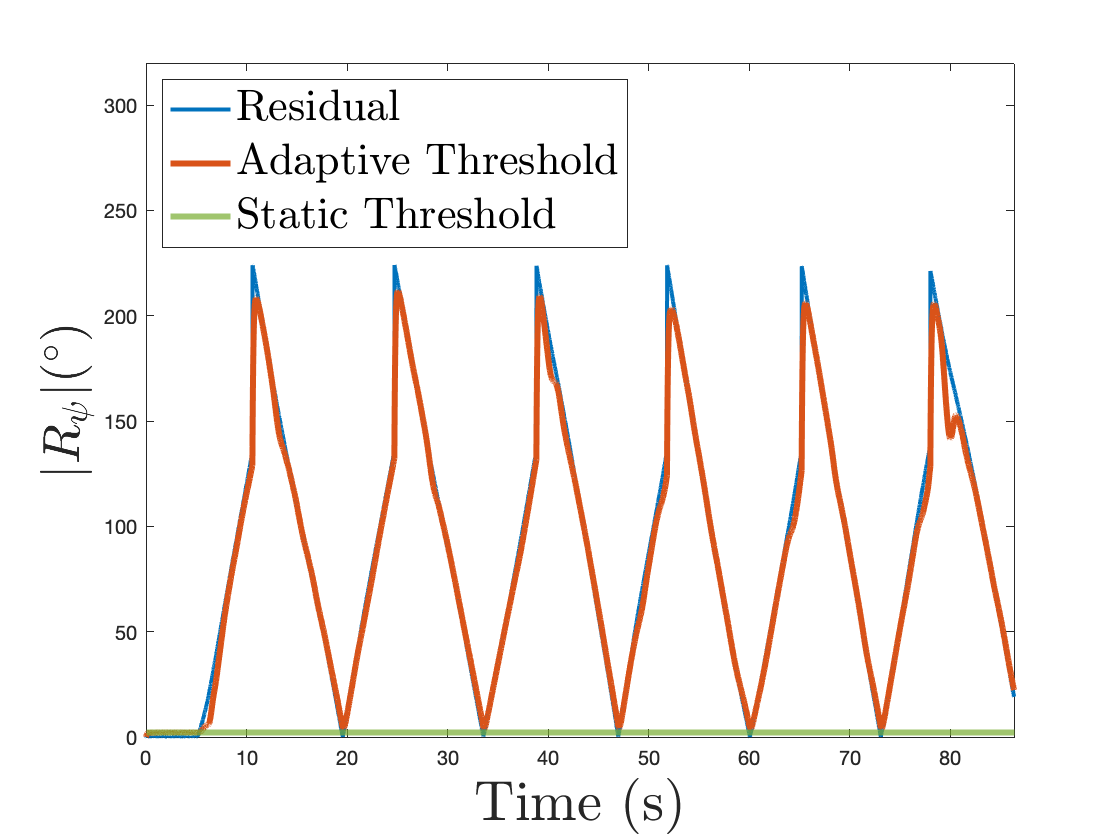}
        \caption{Test case C: heading residual with adaptive threshold in the presence of a motor failure at $t=5$ seconds} \label{fig:1e}
    \end{subfigure} \hspace*{\fill}
    \begin{subfigure}[t]{0.32\textwidth}
        \includegraphics[width=\linewidth]{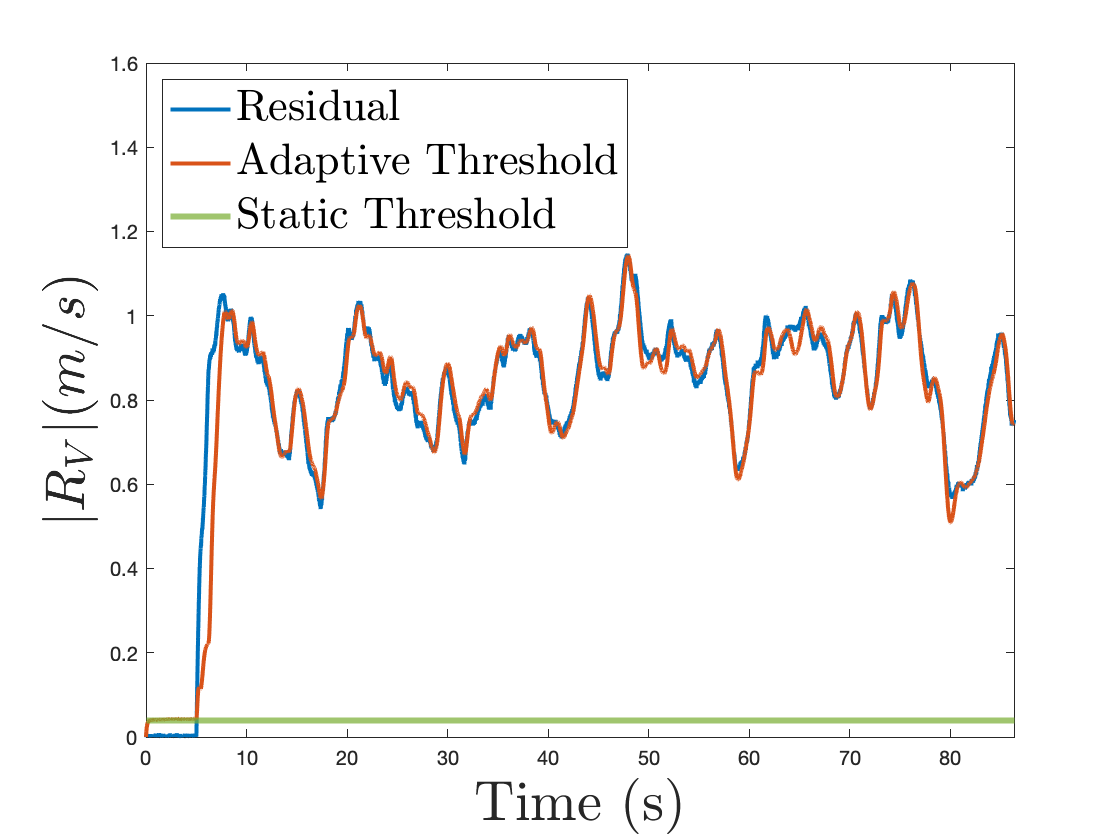}
        \caption{Test case C: velocity residual with adaptive threshold in the presence of a motor failure at $t=5$ seconds} \label{fig:1f}     
    \end{subfigure}

    \caption{Rover path, health monitoring, and adaptive threshold generation for the straight line test scenario} \label{fig:healthAndAdaptiveThresholdsStraight}

    \vspace*{-3mm}
\end{figure*}

Fig. \ref{fig:1d} shows the rover health measurements recorded throughout each of the test cases in the straight line scenario. A clear reduction in health, $H$, is observed at the injection of the fault modes in Test Case B and Test Case C. The residuals ($R_{\psi}$ and $R_V$) for Test Case B are shown in Figs. \ref{fig:1b} and \ref{fig:1c}, respectively. 

\noindent  Here, the heading residual exceeds the adaptive threshold upon fault injection. The adaptive threshold is then able to track the residual as the rover's control system compensates for the fault. The heading and velocity residuals for Test Case C are shown in Figs. \ref{fig:1e} and \ref{fig:1f}, respectively, where it can be seen that both residuals exceed the adaptive thresholds at fault injection. However, the control system is not sufficiently able to compensate for the effects of this fault, and the rover does not return to nominal operation. 
\vspace*{-1mm}
\subsection{Serpentine Path}
\begin{figure*}[t]
    \begin{subfigure}[t]{0.32\textwidth}
        \includegraphics[width=\linewidth]{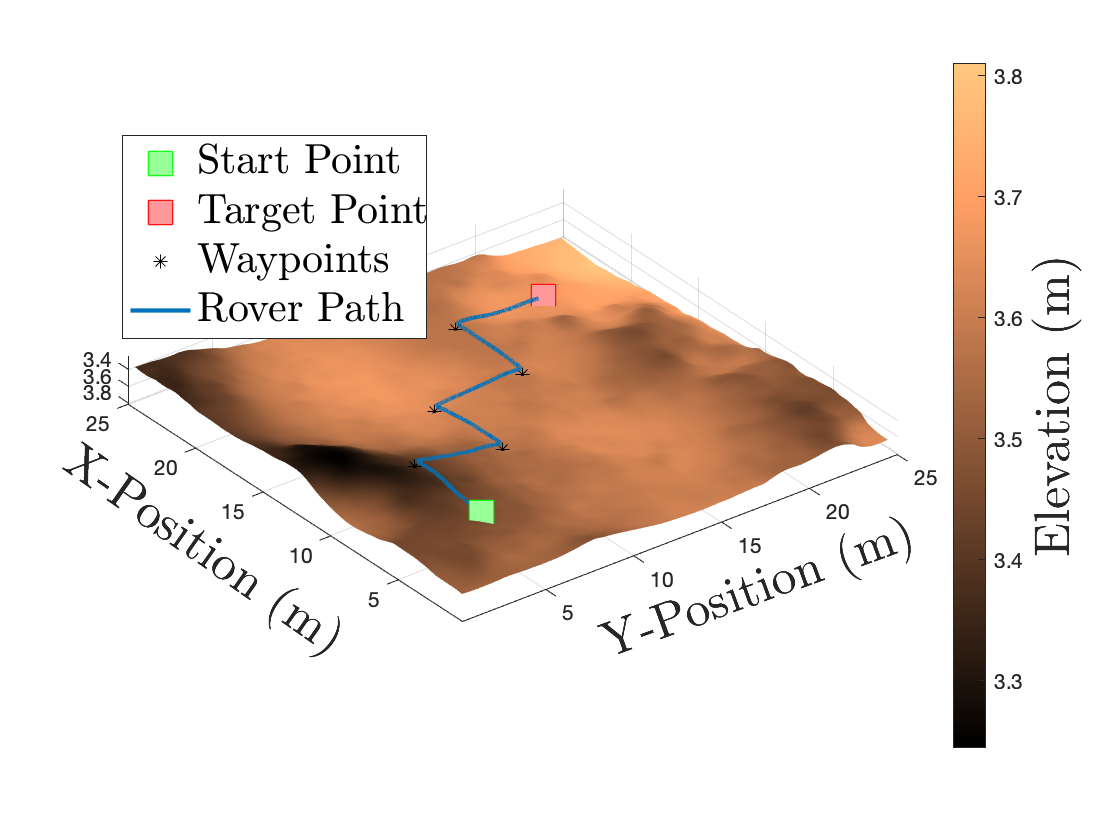}
        \caption{Nominal rover path in the presence of no faults} \label{fig:2a}
    \end{subfigure}\hspace*{\fill}
    \begin{subfigure}[t]{0.32\textwidth}
        \includegraphics[width=\linewidth]{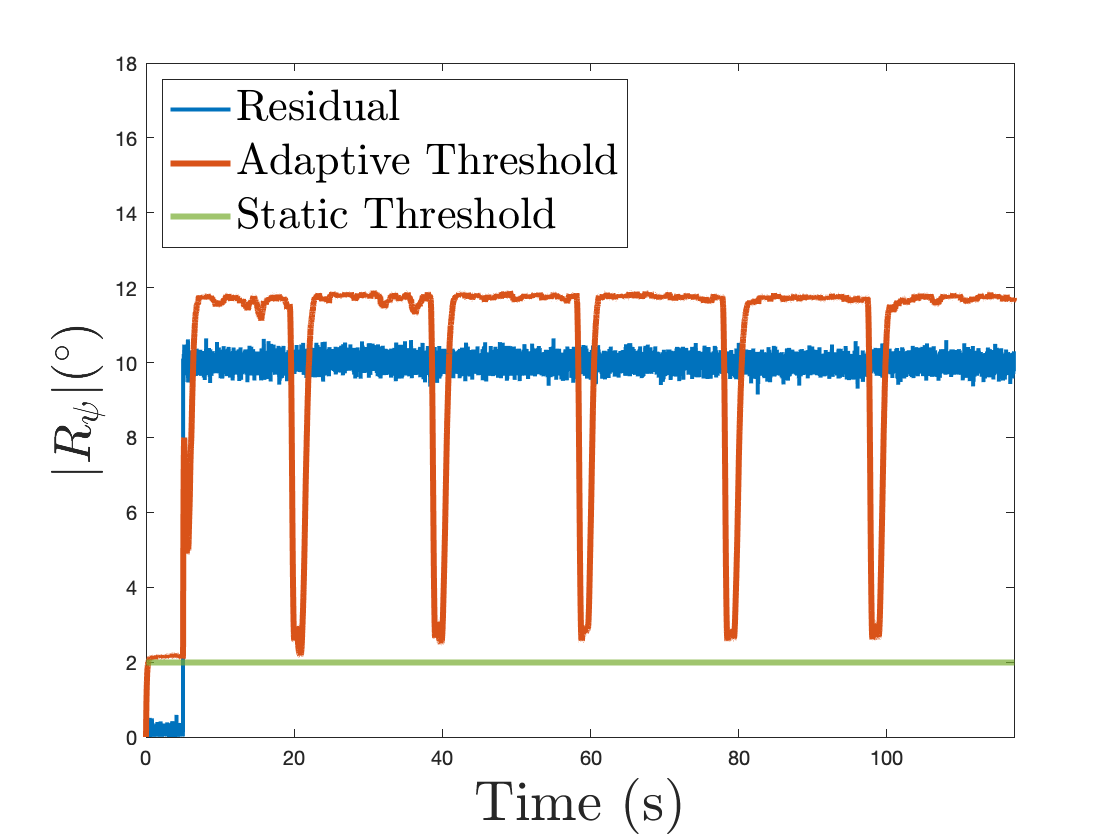}
        \caption{Test case B: heading residual with adaptive threshold in the presence of an abrupt gyroscope offset at $t=5$ seconds} \label{fig:2b}
    \end{subfigure} \hspace*{\fill}
    \begin{subfigure}[t]{0.32\textwidth}
        \includegraphics[width=\linewidth]{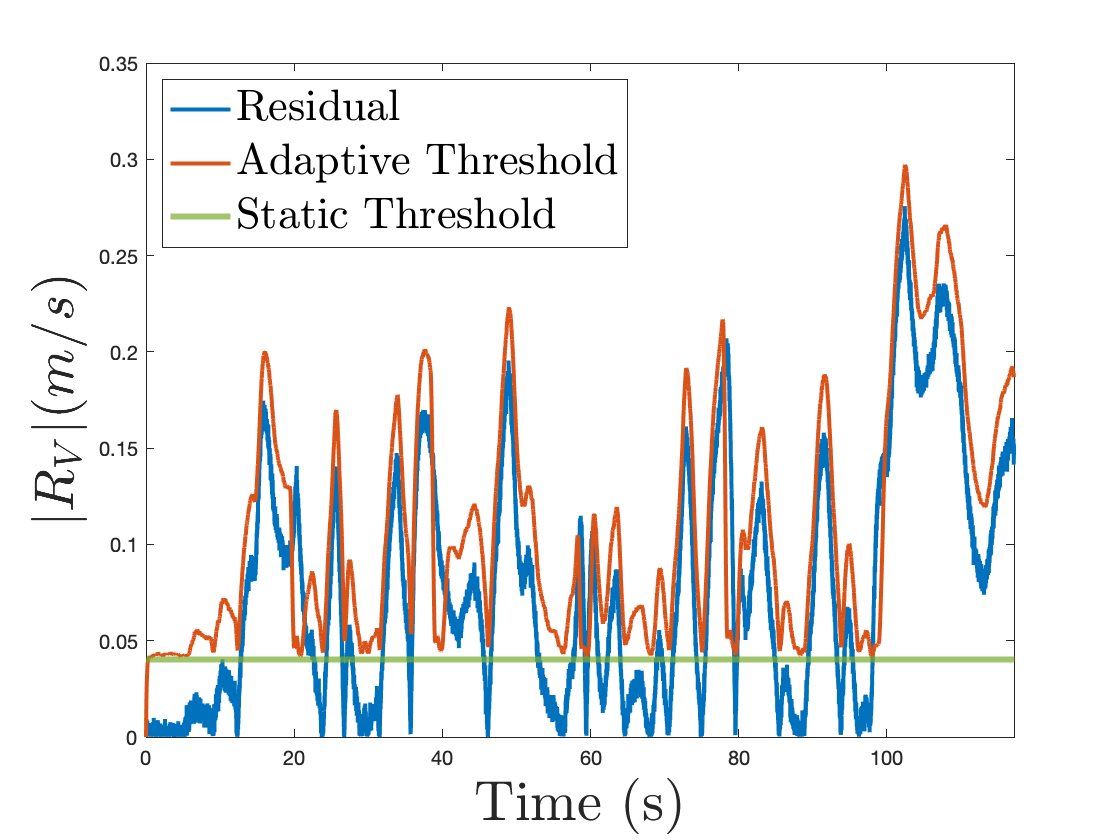}
        \caption{Test case B: velocity residual with adaptive threshold in the presence of an abrupt gyroscope offset at $t=5$ second} \label{fig:2c}     
    \end{subfigure}

    \vspace*{-2mm}

    \medskip
    \begin{subfigure}[t]{0.32\textwidth}
        \includegraphics[width=\linewidth]{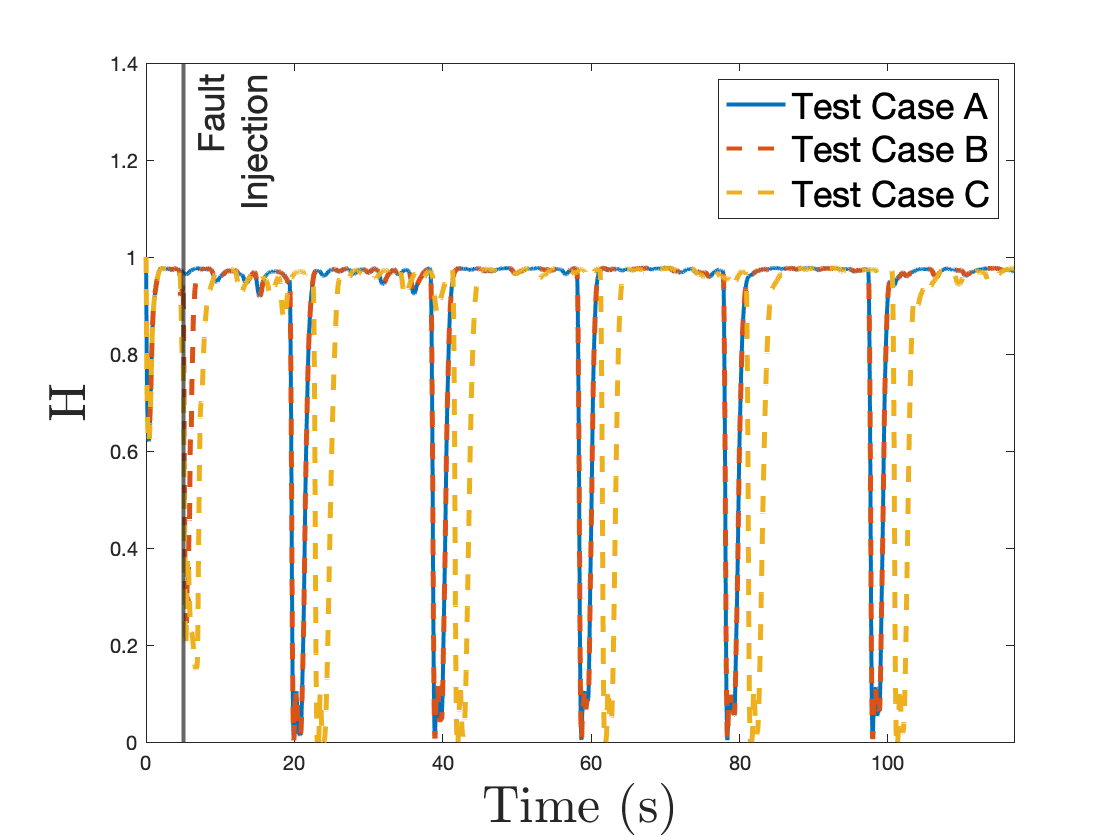}
        \caption{A comparison of the rover health signals for each of the three fault test cases} \label{fig:2d}
    \end{subfigure}\hspace*{\fill}
    \begin{subfigure}[t]{0.32\textwidth}
        \includegraphics[width=\linewidth]{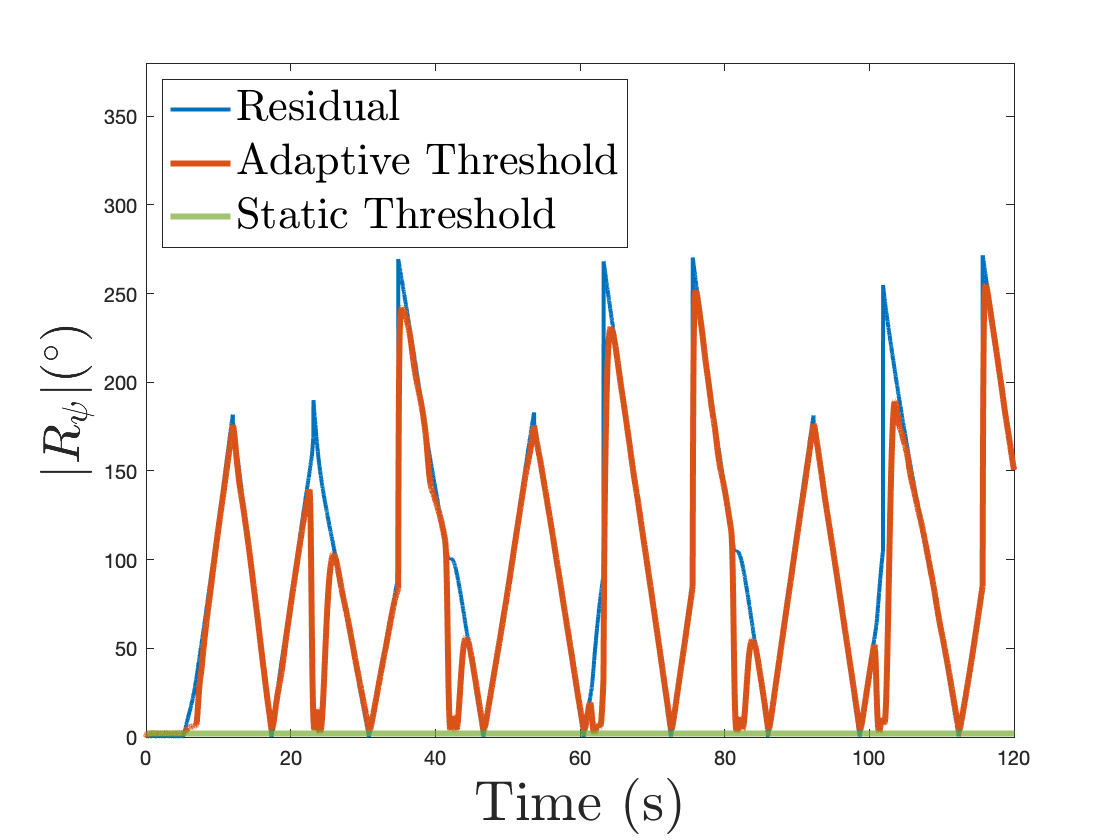}
        \caption{Test case C: heading residual with adaptive threshold in the presence of a motor failure at $t=5$ seconds} \label{fig:2e}
    \end{subfigure} \hspace*{\fill}
    \begin{subfigure}[t]{0.32\textwidth}
        \includegraphics[width=\linewidth]{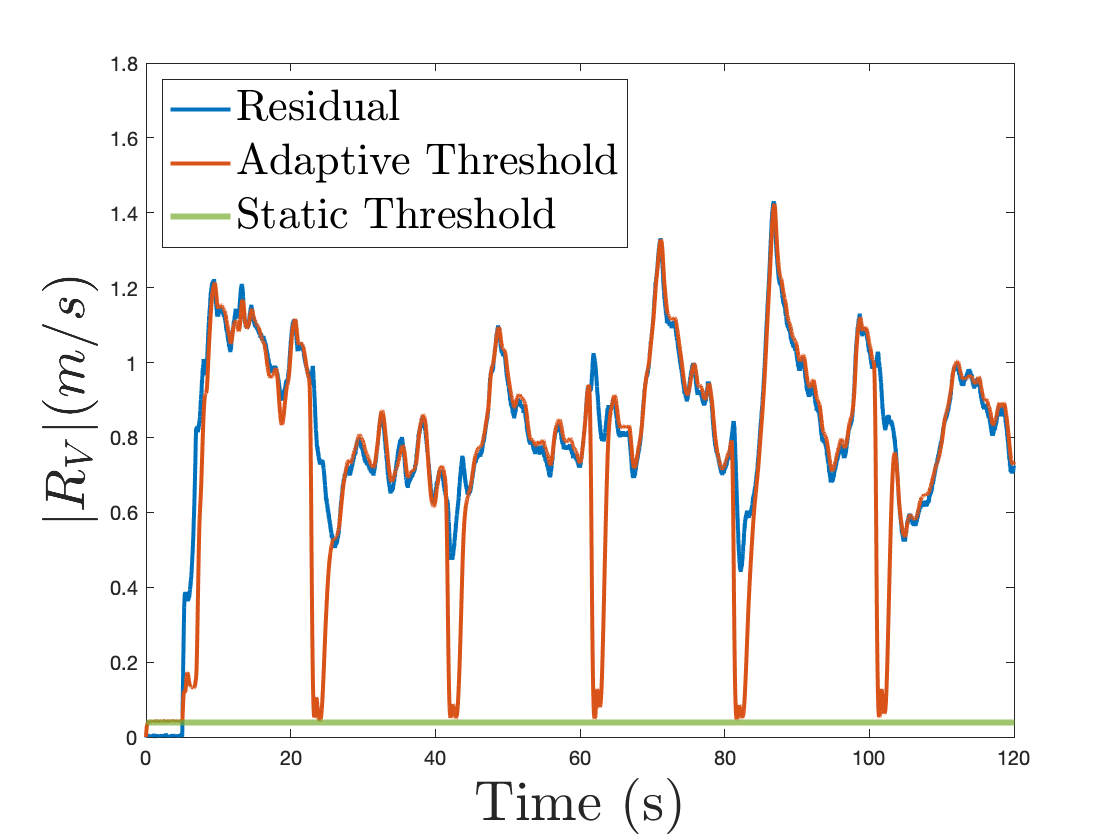}
        \caption{Test case C: velocity residual with adaptive threshold in the presence of a motor failure at $t=5$ seconds} \label{fig:2f}     
    \end{subfigure}

    \caption{Rover path, health monitoring, and adaptive threshold generation for the serpentine path test scenario} \label{fig:healthAndAdaptiveThresholdsSerp}

    \vspace*{-3mm}
\end{figure*}

Fig. \ref{fig:2d} shows the rover health measurements recorded throughout each of the test cases in the serpentine path  scenario. Again, a clear reduction in health, $H$, is observed at the injection of the fault modes in Test Case B and Test Case C. However, the faults injected cause a delay between waypoint collection for the faulty rover and observer. This manifests as a large reduction in $H$. This effect propagates to the generation of the adaptive thresholds, and can be clearly observed in Figs. \ref{fig:2b} and \ref{fig:2f}.

\section{Discussion}\label{sec:discussion}
The health signals produced by the system successfully identify the instantaneous reduction in performance in the rover system due to the two fault modes. The recovery of the rover is captured where the health signal returns to pre-fault levels once the control system compensates for faulty behaviours. Throughout testing, only small fluctuations in measured health are observed as the rover traverses uneven 3D terrain. This suggests that this method is robust to changes in the rover's states (e.g. pitch and roll) when operating within in its nominal limits. However, this method should be made more robust to planned changes in heading, as observed in the serpentine path scenario.

The static thresholds presented here could cause false alarms due to excessive noise or changes in the operational environment. Further, the static thresholds are unable to identify instances where the rover has autonomously recovered from the faulty behaviour. The adaptive fault detection thresholds track both the rover behaviour and its ability to autonomously recover, reducing the likelihood of false positive alarms. This characteristic of the proposed health monitoring and fault detection system is particularly advantageous in the case of an autonomous PER. If the rover is able to compensate for the fault without additional intervention from human operator, autonomous operation can continue, extending the data collection capabilities of a mission. The proposed methods could be extended to help observers diagnose specific faults, allowing other fault recovery methods to be implemented, and further improving the lifespan of a faulty rover.

 \squeezeup \squeezeup
\section{Conclusions}\label{sec:conclusions}

The purpose of this research was to investigate whether health monitoring can be used to inform adaptive thresholds for the purpose of fault detection in PERs. First, a set of rover vitals were established, which quantify the performance degradation of the PER at a given time. Each of these vitals relates to a different aspect of the rover's operational characteristics: acceleration, distance to goal, heading rate, and commanded motor voltage. Nominal rover behaviour was observed to apply probability functions to each vital, where nominal behaviour results in a low probability of performance degradation. A measure of overall rover health was obtained by considering the information entropy associated with the suite of vitals. To enable fault detection, both a faulty rover (representing the real system), and a non-faulty reference rover (acting as an observer) were implemented within simulation, and residuals of heading and velocity were measured. The faulty rover's health measurement was used to inform adaptive thresholds for the purpose of fault detection. Experiments covered three test cases: no fault, gyroscope offset, and motor failure. The proposed fault detection methodology successfully detected the implemented faults.

\end{document}